\documentclass[10pt,twocolumn,letterpaper]{article}

\usepackage[pagenumbers]{cvpr}     
\definecolor{cvprblue}{rgb}{0.21,0.49,0.74}
\usepackage[pagebackref,breaklinks,colorlinks,allcolors=cvprblue]{hyperref}
\PassOptionsToPackage{table}{xcolor}
\usepackage{colortbl}
\usepackage{booktabs}
\usepackage{booktabs}
\usepackage[table]{xcolor}
\usepackage{multirow}
\usepackage{pifont}
\usepackage{multirow}
\usepackage{caption}
\usepackage{newfloat}
\usepackage{listings}
\usepackage{booktabs}
\usepackage{multirow}
\usepackage{array}
\usepackage{cuted}
\usepackage{caption}
\usepackage{cuted}
\usepackage{capt-of} 
\usepackage{makecell}
\usepackage{placeins}
\usepackage{bibentry}
\usepackage{pifont}
\usepackage{amssymb}
\usepackage{subcaption}

\usepackage[utf8]{inputenc}
\usepackage{url}
\usepackage{booktabs}
\usepackage{amssymb}
\usepackage{bbding}
\usepackage{pifont}
\usepackage{wasysym}
\usepackage{utfsym}
\usepackage{fontawesome}
\def\paperID{18388} 
\def\confName{CVPR}
\def\confYear{2026}

\title{Plug-and-Play Homeostatic Spark: Zero-Cost Acceleration for SNN Training Across Paradigms}

\author{
Rui Chen\\
Peking University, Beijing, China\\
{\tt\small chenruiii@stu.pku.edu.cn}
\and
Xingyu Chen\\
C9, Dayue Information Technology Park, Haidian District, Beijing, China\\
{\tt\small chenxy.sean@gmail.com}
\and
Yaoqing Hu\\
C9, Dayue Information Technology Park, Haidian District, Beijing, China\\
{\tt\small huyaoqing@bjzgca.edu.cn}
\and
Shihan Kong\\
Peking University, Beijing, China\\
{\tt\small kongshihan@pku.edu.cn}
\and
Zhiheng Wu\\
Baidu Inc., Beijing 100085, China\\
{\tt\small wzh404.ai@gmail.com}
\and
Junzhi Yu\\
Peking University, Beijing, China\\
{\tt\small yujunzhi@pku.edu.cn}
}

\begin{document}
\maketitle

\begin{strip}
  \centering
  \includegraphics[width=1.0\textwidth]{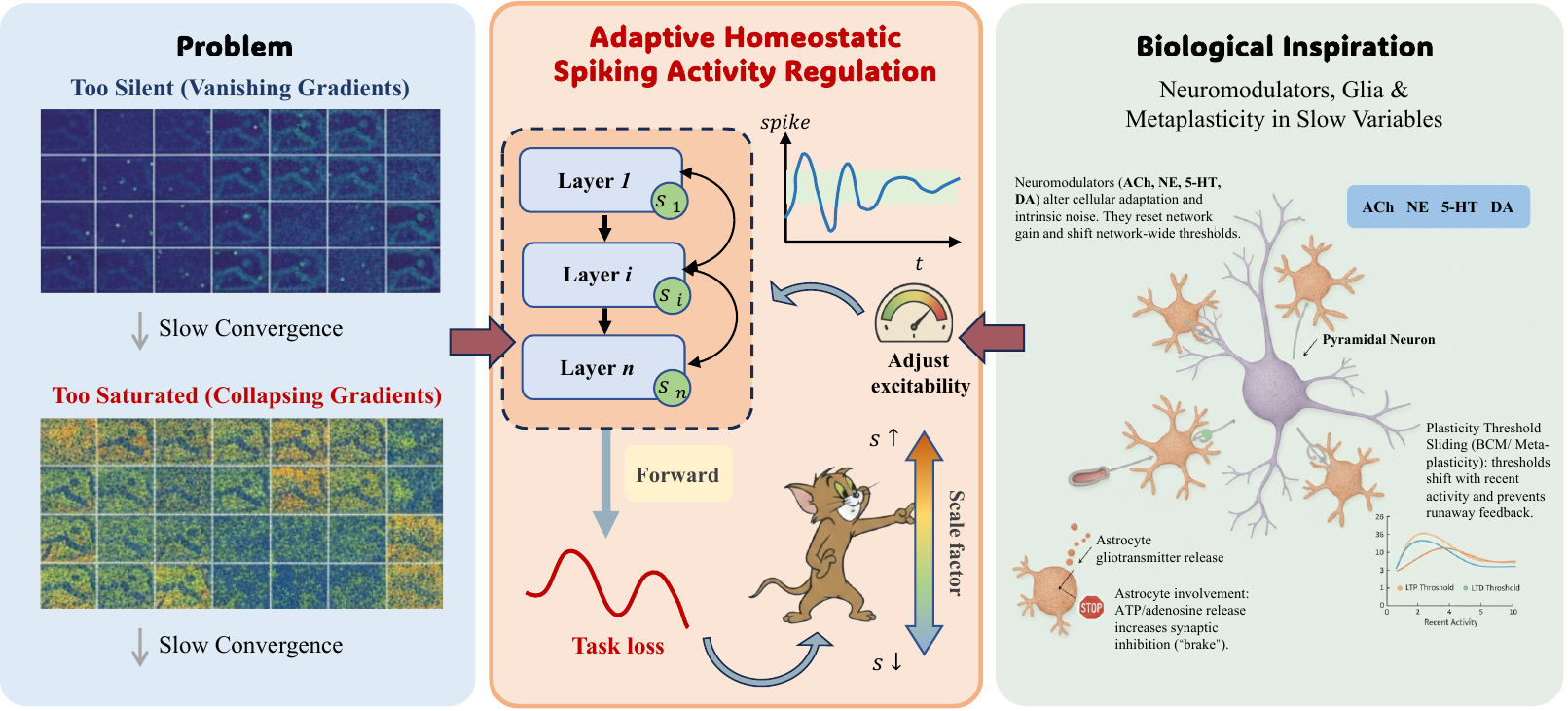}
  \captionof{figure}{Overview of current problem, biological inspiration and our method.}
  \label{fig:teaser}
\end{strip}
 
\begin{abstract}

Spiking neural networks offer event driven computation, sparse activation, and hardware efficiency, yet training often converges slowly and lacks stability. We present Adaptive Homeostatic Spiking Activity Regulation (AHSAR), an extremely simple plug in and training paradigm agnostic method that stabilizes optimization and accelerates convergence without changing the model architecture, loss, or gradients. AHSAR introduces no trainable parameters. It maintains a per layer homeostatic state during the forward pass, maps centered firing rate deviations to threshold scales through a bounded nonlinearity, uses lightweight cross layer diffusion to avoid sharp imbalance, and applies a slow across epoch global gain that combines validation progress with activity energy to tune the operating point. The computational cost is negligible. Across diverse training methods, SNN architectures of different depths, widths, and temporal steps, and both RGB and DVS datasets, AHSAR consistently improves strong baselines and enhances out of distribution robustness. These results indicate that keeping layer activity within a moderate band is a simple and effective principle for scalable and efficient SNN training.
\end{abstract}    
\section{Introduction}
\label{sec:intro}

Spiking neural networks offer event driven computation, sparse activation, and hardware friendliness, which makes them attractive for low power perception and control \cite{roy2019towards}. However, training efficiency remains a major bottleneck. State of the art systems often require many epochs to reach competitive accuracy, and transfer across architectures and datasets is fragile \cite{neftci2019surrogate,li2021differentiable}. Current practice follows two main paths. The first uses backpropagation through time with surrogate gradients to handle the discontinuous spike function. The second uses rate based training such as rate propagation that decouples temporal dynamics and optimizes a differentiable rate representation \cite{yu2024advancing}. Both paths benefit from careful threshold schedules and normalization, yet they still show slow starts, oscillatory loss, and sensitivity to hyperparameters \cite{kim2021revisiting,duan2022temporal}.

To improve stability, prior work often changes neuron dynamics or inserts auxiliary modules. Examples include adaptive thresholds driven by running activity, synaptic or membrane normalization, and post training conversion pipelines that search layer wise thresholds. These approaches tend to rely on architecture specific heuristics, add nontrivial computation, or deliver gains that hold only within one training paradigm. This reality motivates a minimal mechanism that does not change model structure, that is compatible with both backpropagation through time and rate based training, and that adaptively regulates layer excitability during learning.

Neuroscience offers a useful principle. Cortical networks maintain stable activity through intrinsic plasticity, synaptic scaling, and the balance between excitation and inhibition \cite{turrigiano2004homeostatic,vogels2011inhibitory}. Neurons adjust their excitability so that firing rates stay within an efficient range \cite{turrigiano2008self}. Inspired by this view, we hypothesize that each layer in a spiking network has a productive activity window. When activity falls below this window, neurons remain silent and gradient flow weakens. When activity exceeds this window, layers saturate, surrogate gradients flatten, and statistics become unstable. Keeping layers inside this window should accelerate optimization.

A large body of studies shows that healthy neural circuits regulate firing activity within a safe operating range and return to a set point after perturbations through homeostatic control acting at synapses and intrinsic excitability \cite{marder2006variability}. When firing is chronically high, networks pay a large metabolic cost for spike generation and face risks such as excitotoxic damage \cite{attwell2001energy,lennie2003cost}. When firing is chronically low, coding capacity and plasticity deteriorate, and circuits fail to maintain useful representations. Motivated by these observations, we aim to keep layer wise firing in a moderate band during training. To verify its necessity, we manually adjusted firing rates across different network architectures and training methods and ran multiple comparative experiments. The results show that for spiking neural networks on datasets in both RGB and DVS modalities, both overly high and overly low firing rates significantly slow optimization and delay convergence.

Based on the above motivation, we propose Adaptive Homeostatic Spiking Activity Regulation (AHSAR). With minimal parameter and runtime overhead, AHSAR guides the spiking activity of each layer into a task-adaptive moderate range. The method is designed as a plug-and-play mechanism: it can be directly attached to a standard spiking network without modifying the original structure or training pipeline, and can be integrated into existing implementations with only minor adaptations. AHSAR operates through a three-stage control loop: it first estimates layer-wise firing rates over physical or virtual time steps, then applies a reaction–diffusion process to compute adaptive threshold scales, and finally incorporates a global gain modulated by validation performance and network energy. A reaction term drives each layer’s excitability state based on its firing rate deviation from a dynamic reference. We incorporate a lightweight cross-layer diffusion term to weakly couple adjacent layers. This enables dual-timescale regulation—fast balancing of inter-layer activity and slow adaptation of the global firing regime. The mechanism is agnostic to the training paradigm, requires no changes to the network architecture, and introduces no extra trainable parameters.

We evaluate AHSAR under both training paradigms. For backpropagation through time it accelerates early optimization and reduces loss oscillation without changing the unrolled dynamics. For rate based training it smooths rate statistics and improves the stability of surrogate normalization. Experiments across large scale datasets and diverse architectures show consistent gains, and the method remains robust across changes in depth, width, and temporal steps. Our main contributions are as follows:	
\begin{itemize}
    \item Through biologically grounded modeling informed by neuroscience and an extensive suite of experiments, we establish a robust link between layer-wise spiking activity and training convergence behavior.
    \item We introduce AHSAR, a plug-and-play method for spiking neural networks. It adaptively modulates neuronal excitability using layer-wise firing rates and enables cross-layer coordination through lightweight diffusion. This approach ensures stable and efficient training while maintaining compatibility with various training paradigms, standard SNN architectures, and diverse input modalities. The method requires no additional training parameters and introduces minimal computational overhead.
    \item We conduct comprehensive evaluations against strong baselines and analyze convergence, out-of-distribution robustness, and energy consumption, demonstrating that AHSAR delivers faster convergence at extremely low complexity while keeping the training pipeline simple.
\end{itemize}

\section{Related Work}
\label{sec:formatting}

Spiking neural networks are trained along two routes. The first uses surrogate gradients for direct backpropagation through time \cite{shrestha2018slayer,lee2016training,lee2020enabling}. Although surrogate gradient BPTT reaches competitive accuracy, it is sensitive to layerwise spiking activity and temporal statistics, which undermines stability \cite{meng2022training,meng2023towards,chen2023training,guo2024enof}. This has motivated spike oriented normalization such as BNTT and membrane potential normalization before the spike nonlinearity to smooth dynamics and reduce loss oscillations. These techniques complement threshold scheduling and membrane or weight normalization but often require extra tuning for specific architectures or time steps.

A widespread practical assumption is to increase firing rates to obtain faster information transfer and fewer inference timesteps. Early ANN to SNN conversion studies often pushed certain layers to high spike densities via max or channel normalization or pre charged membrane potentials \cite{diehl2015fast,cao2015spiking,bu2022optimized,ho2021tcl}. While this helps at short horizons, our experiments show that persistently high firing in deep or upper layers can slow optimization and increase gradient noise, yielding active yet slowly converging training. Maximizing spike rate therefore has a cost and must be balanced with separability, timestep budgets and energy constraints.

A parallel approach avoids non differentiability during learning by first training an ANN and converting it to an SNN through threshold calibration and rescaling or by using rate based approximations that compress temporal dependencies \cite{jiang2023unified}. Classical conversion improves the accuracy latency trade off. More recent work refines thresholds and initializations to further narrow the ANN SNN gap and shorten inference . Rate based backpropagation leverages average firing rates to reduce memory and wall clock time while approaching BPTT accuracy. All these routes still benefit from mechanisms that keep layer excitability within an efficient window. To avoid both under firing that dilutes information and over driving that slows convergence, a bidirectional controller is desirable \cite{zenke2017temporal,ruggiero2021mitochondria}.

From an engineering viewpoint, normalizing postsynaptic potentials or time dependent batch parameters can clamp spike rates and stabilize gradients in deeper SNNs. These controls are often tied to specific insertion points or require per timestep parameters, complicating deployment across architectures and temporal scales. Introducing a target activity band addresses this by suppressing excitability when a layer stays above the upper bound and by increasing excitability when it remains below the lower bound, for example via threshold reduction or increased membrane integration gain. This covers layers deliberately driven high for short latency as well as layers fed by sparse event inputs.

Neuroscience supports this view. Cortical circuits maintain firing within a homeostatic range through synaptic scaling, intrinsic excitability regulation and inhibitory plasticity that preserve excitatory inhibitory balance. Maintaining moderate activity is therefore a core organizational principle. Recently, BDETT \cite{ding2022biologically} proposed a bio-inspired dynamic energy-temporal threshold scheme. However, this method relies on constants fitted from specific nervous system data and lacks generality across diverse SNN architectures and training paradigms, thereby altering the original training paradigm.

Neuromodulators are slow variables that gate excitability and energy use in both directions. Adenosine via A1 receptors suppresses glutamatergic transmission and hyperpolarizes neurons, with extracellular levels accumulating during wakefulness and high activity to provide negative feedback from metabolism to excitability \cite{dunwiddie2001role}. In contrast, acetylcholine and norepinephrine increase cortical firing rates and input output gain during attentive states, and dopamine and orexin can elevate firing in specific circuits \cite{picciotto2012acetylcholine,aston2005integrative}. An SNN regulator inspired by neuromodulation should therefore support both down regulation of excessive activity and up regulation when activity is too low.

SNNs are evaluated on frame based RGB datasets and event based neuromorphic benchmarks \cite{orchard2015converting,amir2017low}. CIFAR10 DVS emphasizes low latency and sparse processing, while larger variants such as N ImageNet and ES ImageNet challenge both conversion and direct training at realistic class granularity. A homeostatic mechanism that keeps layers within an appropriate range is broadly useful across modalities and it serves the single goal of operating where gradients propagate, energy is affordable and temporal structure is preserved \cite{abbott2000synaptic}.
\section{Spiking Activity Analysis}
\label{sec:formatting}

\begin{figure*}[t]
\centering
\includegraphics[width=1.0\textwidth]{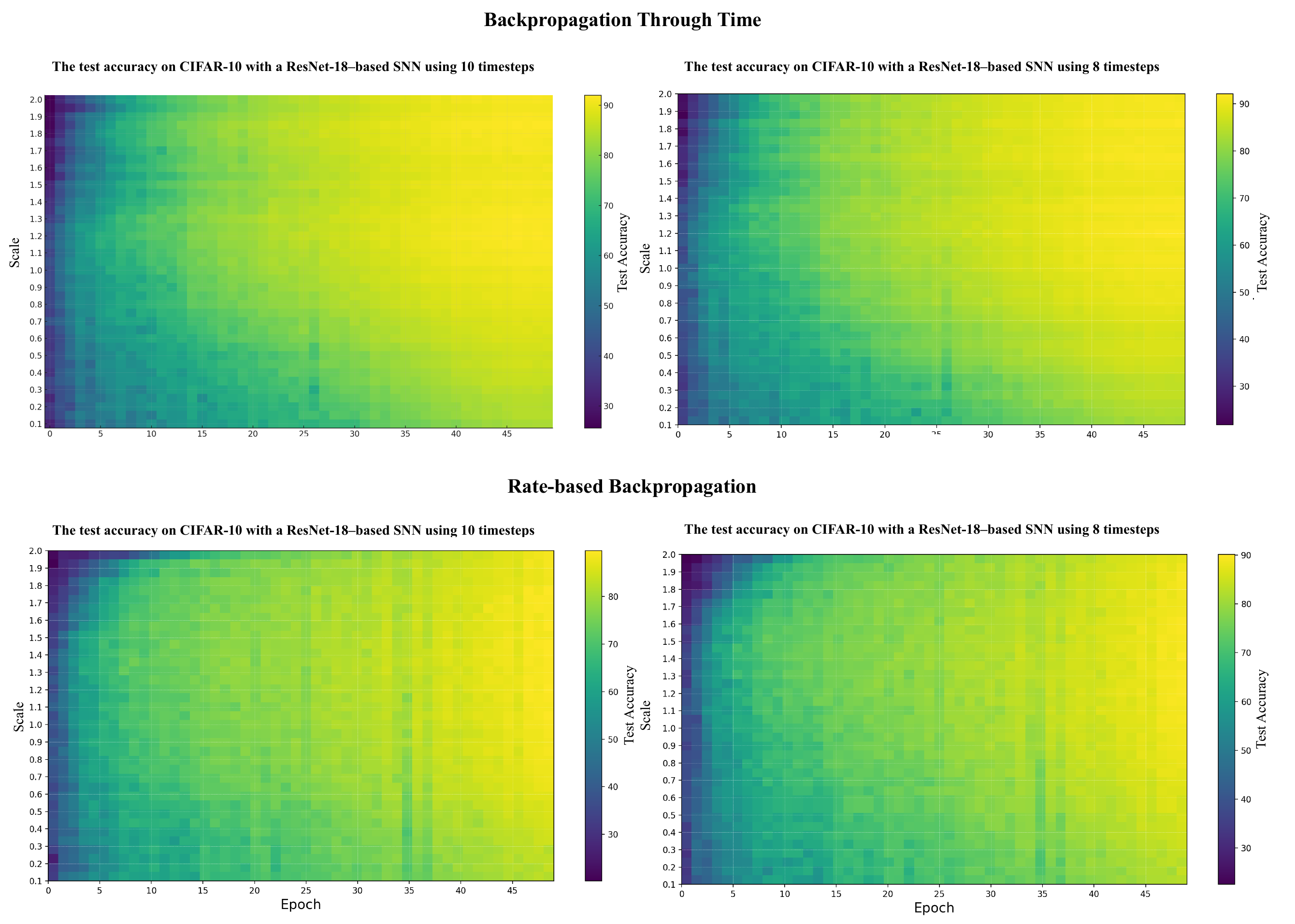} 
\caption{The impact of different scaling factors $S$ on model convergence under different training methods.}

\label{fig2}
\end{figure*}

Neuroscience indicates that cortex maintains excitatory-inhibitory balance and homeostatic plasticity, which motivates modeling the subthreshold voltage as a noisy leaky integrate-and-fire diffusion that relaxes toward a mean drive $\mu$ with time constant $\tau$ and noise scale $\sigma$
$$dU_{t}=-\frac{1}{\tau}(U_{t}-\mu)dt+\sigma dW_{t}.$$
where $U_{t}$ is the subthreshold membrane potential, $W_{t}$ is a standard Wiener process, $\mu$ is the mean input, $\tau$ is the leak time constant, and $\sigma$ sets the fluctuation magnitude. A spike is emitted when the voltage crosses a fixed threshold $\theta$, and under stationarity each step behaves like a binary event with rate $r\equiv\operatorname{Pr}(U_{t}\geq\theta)$ that decreases as $\theta$ increases. During training a smooth surrogate $\psi(U-\theta)$ replaces the hard trigger, which creates a gradient gate $g=\psi'(U-\theta)$ concentrated near the threshold, and taking expectation under the stationary law yields an average gate that aligns with the stationary density at the boundary, as follows:

\vspace{-10pt}

$$G(\theta)=\mathbb{E}\left[\psi^{\prime}(U-\theta)\right]=\left(\psi^{*} p_{U}\right)(\theta)\approx p_{U}(\theta)=-\frac{\partial r}{\partial\theta},$$

where $G(\theta)$ is the mean gate, $p_{U}$ is the stationary density of $U$, and $*$ denotes convolution. From an information view, extremely low activity weakens signal propagation and leads to vanishing gradients, whereas extremely high activity reduces selectivity and increases correlation and gradient noise while energy usage grows with activity. From an optimization and dynamics view, the effective sensitivity through a layer is shaped by the gate and the weights, which is given by:

$$J_{\text{eff}} \approx \operatorname{diag}(G(\theta)) W,$$

in which W is the synaptic weight matrix and $\operatorname{diag}(\cdot)$ places its entries on the diagonal. These complementary effects imply an intermediate activity band that balances information flow, separability, gradient stability, and computational or metabolic cost.

To further validate the influence of the threshold on spiking activity and learnability, we conduct a set of strictly controlled experiments: in each run, all settings are kept identical except for the threshold. Concretely, we multiply the firing threshold of every LIF layer by a global scaling factor $s$, keep this factor fixed throughout both training and evaluation, and share the same $s$ across all layers. We sweep $s$ from 0.10 to 2.00 with a step of 0.05 and compare convergence speed, stability, and final accuracy under different threshold scales. This design isolates the effect of threshold scaling on training dynamics and generalization without introducing additional confounders. The experimental results in Figure \ref{fig2} validate our hypothesis and provide empirical support. We further observe that, for each training paradigm, there exists a range of threshold values within which the model converges faster.
\section{Method}
\label{sec:formatting}

\begin{figure*}[t]
\centering
\includegraphics[width=1.0\textwidth]{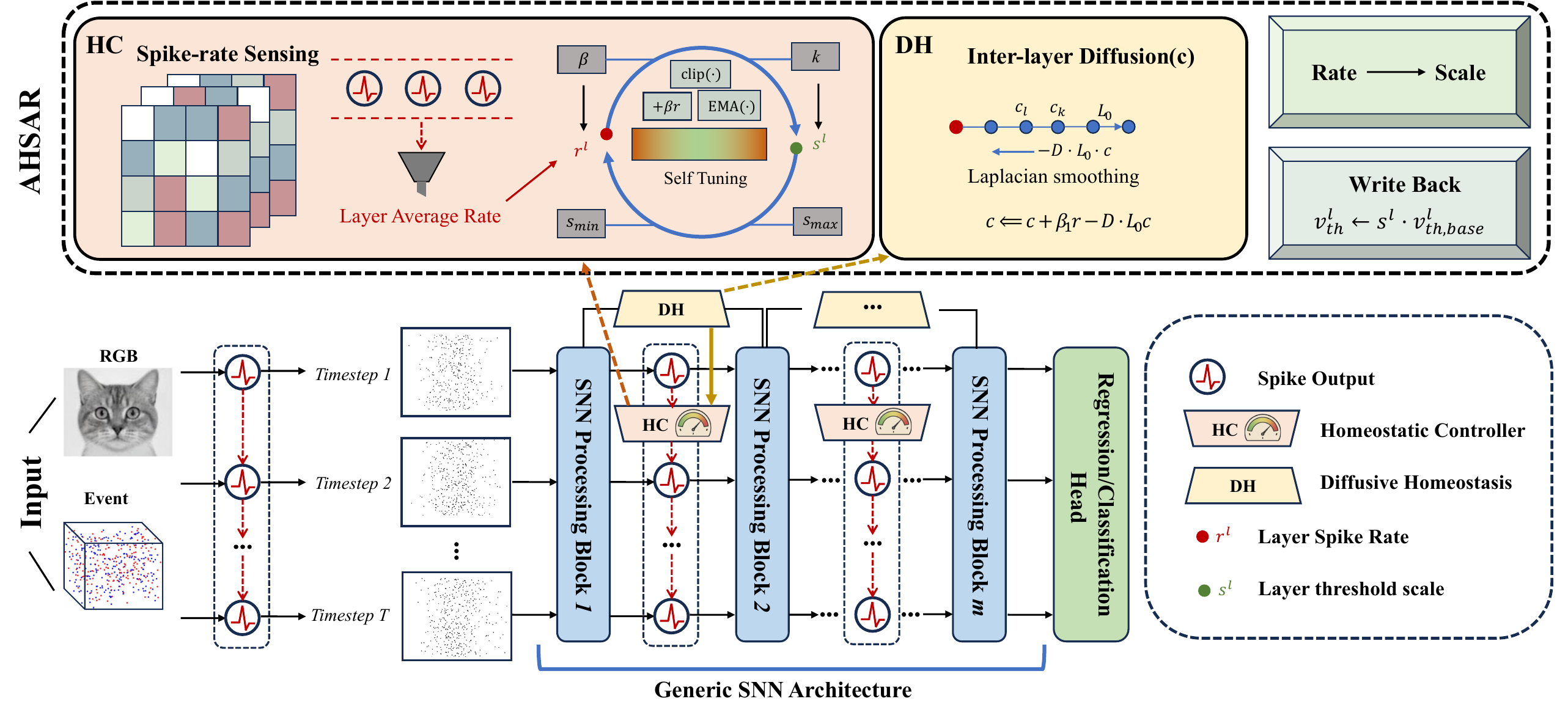} 
\caption{AHSAR overview. The Homeostatic controller senses layer-average spike rates, smooths and centers them. A diffusive term propagates corrections across neighboring layers via Laplacian smoothing, and the scaled thresholds are written back to each layer. The modules plug into a generic SNN over timesteps without changing the original forward/backward.}

\label{fig3}
\end{figure*}
Training SNNs remains sensitive to the distribution of spike activity across layers and across time, which often leads to slow convergence and brittle transfer across architectures. We introduce AHSAR, an adaptive homeostatic spike-aware regulation scheme that treats the layer-wise spike rate as a primary feedback signal and converts it into threshold modulation. The method is designed as a controller that operates on top of an existing simulator and optimizer and does not alter the training loss or the gradient computation. It therefore acts as a plug-in component that can be attached to standard surrogate gradient training or rate-based objectives. The regulator introduces no trainable parameters. It maintains per-layer states during a forward run and writes bounded threshold scales back to layers without adding variables to the optimization, which makes the mechanism plug-in and architecture-agnostic.

AHSAR operates on two time scales. The inner loop runs within an epoch and redistributes excitability across depth using centered rate deviations and a smooth bounded mapping to thresholds. Overactive layers receive larger thresholds and calm down. Underactive layers receive smaller thresholds and participate more. Diffusion spreads adjustments to neighboring layers and avoids sharp imbalances. The loop has a stable fixed point where deviations vanish and layers share work. The outer loop runs across epochs and modulates a single global gain using validation progress and activity energy. When progress stalls while energy is high, the gain increases and suppresses spikes. When progress improves while energy is low, the gain decreases and permits more spikes. The fixed point of this rule is a regime where changing the gain no longer improves validation loss and energy stays near its trend. Together, the two loops move the network away from unproductive extremes and toward a productive band without a hand-picked target rate.

The proposed controller follows a structured three-stage pipeline. First, we build a unified rate-sensing module that aggregates binary spikes into scale-free statistics and aligns models with explicit timesteps and models in rate mode through a notion of virtual control ticks. Second, we introduce a layer-wise reaction-diffusion dynamics that evolves a low-dimensional homeostatic state per layer and converts this state into threshold scales through a smooth saturating mapping. Third, we integrate an epoch-level global gain that reacts to long-range trends in validation performance and network energy and modulates all thresholds in a coherent manner. The combination yields a two-time-scale adaptation process. Fast dynamics redistribute excitability across layers within an epoch, and slow dynamics shift the global operating point across epochs.

We define the control ticks $\hat{T}$ based on the training mode: for step mode $s$, $\hat{T} = T$ (number of physical timesteps); for rate mode $m$, $\hat{T} = K$ (number of conceptual stages, e.g., residual blocks). The controller updates once per control tick, ensuring consistent rate sensing and regulation for both paradigms.

\subsection{Rate sensing across physical and virtual time}

The first component of the method aims to map raw binary spikes into stable layer wise statistics that can be used as feedback signals. For each control tick $\tau\in\{1,\ldots,\hat{T}\}$ and each LIF based layer $l$, we define an instantaneous spike rate that averages over batch and neuron dimensions. In step mode $s$ the simulator produces spikes $z^{l}_{b,i,t}$ at physical timestep $t\in\{1,\ldots,T\}$. For a given optimization step we define the rate of layer $l$ as follows:
$$r^{l}=\frac{1}{|\mathcal{B}| T|\Omega^{l}|}\sum_{b\in\mathcal{B}}\sum_{t=1}^{T}\sum_{i\in\Omega^{l}}z^{l}_{b,i,t}$$
where $\mathcal{B}$ denotes the mini batch index set, $|\Omega^{l}|$ is the number of neurons in layer $l$ and $z^{l}_{b,i,t}$ indicates whether neuron $i$ in layer $l$ fires for sample $b$ at time $t$. In rate mode $m$ the simulator exposes one effective activation per layer per forward pass. We align this setting with the control ticks by aggregating activations over predefined blocks. For a block that spans a subset of layers $\mathcal{L}_{\kappa}$ we define a virtual rate as follows:
$$r^{\kappa}=\frac{1}{|\mathcal{B}||\mathcal{L}_{\kappa}|}\sum_{b\in\mathcal{B}}\sum_{l\in\mathcal{L}_{\kappa}}\phi\left(u^{l}_{b}\right)$$
where $u^{l}_{b}$ is the membrane potential proxy or pre activation of layer $l$ for sample $b$ and $\phi$ is a differentiable surrogate nonlinearity that estimates the expected spike probability. The virtual rate $r^{\kappa}$ acts as a control signal for the block and we attach one homeostatic state to each block in this case.

Instantaneous rates are typically noisy due to variation across batches and early training dynamics. To obtain a smoother signal the controller maintains an exponential moving average per controlled unit. For each layer or block $u$ we maintain a smoothed rate $\tilde{r}^{u}_{t}$ at optimization step $t$ through the following update:
$$\tilde{r}^{u}_{t}=\left(1-m_{r}\right)\tilde{r}^{u}_{t-1}+m_{r}\,r^{u}_{t}$$
where $r^{u}_{t}$ is the current batch estimate and $m_{r}\in(0,1]$ is a rate momentum parameter. The smoothed rate is initialized from the first batch statistics and is updated once per optimization step, aggregated across all control ticks in that step. This construction is independent of the underlying training loss, only requires reading spike counts or surrogate activations that are already produced by the forward simulation, and introduces a negligible memory overhead of one scalar per controlled unit. It forms the primary measurement signal for the subsequent reaction diffusion dynamics. The centering used later will act on $\log\tilde{r}$ so the signal becomes scale free and comparable across layers and across datasets, which is essential for a stable closed loop.

\subsection{Layer wise reaction diffusion homeostasis}

Given the smoothed rates we define a minimal dynamical state that tracks the excitability of each layer and propagates information across depth. For each controlled unit $u$ we introduce a scalar homeostatic state $c_{t,\tau}^{u}$ indexed by optimization step $t$ and control tick $\tau$. Intuitively a unit that fires more should accumulate a larger state which in turn increases its effective threshold and decreases future spike activity. Units that are adjacent in depth should influence each other so that a highly active layer can attenuate its neighbors and a silent layer can borrow excitability from nearby ones.

The reaction diffusion update for the state vector $\mathbf{c}_{t,\tau}=\left[c^{1}_{t,\tau},\ldots,c^{L}_{t,\tau}\right]^{\top}\in\mathbb{R}^{L}$ is given as follows:
$$\mathbf{c}_{t,\tau+1}=\mathbf{c}_{t,\tau}+\beta(\log\tilde{\mathbf{r}}_{t,\tau}-\bar{h}_{t}\mathbf{1})-\gamma\left(\mathbf{c}_{t,\tau}-\mathbf{c}_{0}\right)-D L\mathbf{c}_{t,\tau}$$
where $\tilde{\mathbf{r}}_{t,\tau}$ collects the smoothed rates at tick $\tau$, $\beta>0$ is a reaction gain, $\bar{h}_{t}$ is a scalar reference computed as the mean of $\log\tilde{\mathbf{r}}_{t,\tau}$ across units, $\gamma>0$ is a relaxation gain, $\mathbf{c}_{0}$ is a learnable or fixed baseline state, $D>0$ is the diffusion strength and $L$ is the discrete Laplacian matrix of the chain that connects neighboring layers. The term in $\beta$ reacts to deviations of the log rate from the global reference and increases the state for over active units while decreasing it for under active ones. The relaxation term in $\gamma$ prevents unbounded drift by pulling the state back to $\mathbf{c}_{0}$ over long horizons. The diffusion term in $D$ enforces smoothness across depth by penalizing sharp changes in $c_{t,\tau}^{u}$ between adjacent units.

The Laplacian $L$ is tridiagonal with entries defined as follows:
\[
L_{ij} =
\begin{cases}
2, & i=j \text{ and } 1<i<L,\\
1, & i=j\in\{1, L\},\\
-1, & |i-j|=1,\\
0, & \text{otherwise}.
\end{cases}
\]

which implements a second order finite difference with natural boundary conditions. The update is first order and can be evaluated with linear cost in the number of layers, which keeps it compatible with deep networks. In step mode $s$ the controller steps through $\tau=1,\ldots, T$ within each optimization step. In rate mode $m$ the same equations are applied across virtual micro steps where each tick corresponds to a block in the depth partition. This ensures that the reaction diffusion dynamics see a comparable notion of temporal progression in both training paradigms.

To influence the neuron dynamics we map the homeostatic state to a positive threshold scale through a smooth saturating nonlinearity. For each unit $u$ we define the scale as:
$$s^{u}_{t,\tau}=s_{\text{base}}\left(1+\alpha \tanh \left(\kappa c^{u}_{t,\tau}\right)\right)$$
where $s_{\text{base}}$ is a reference scale, $\alpha>0$ controls the maximum relative deviation from the base scale and $\kappa>0$ shapes the sensitivity of the mapping. The hyperbolic tangent ensures that large positive or negative states do not cause unbounded changes in threshold and that the mapping is monotone in the state. This produces a per layer scale that encodes the current excitability in a way that is smoothly differentiable with respect to the state, although we treat the controller as a detached process and do not backpropagate through it. The inner loop thus forms a negative feedback with damping and a bounded actuator. It converges to a fixed point where rate deviations fade and layers operate in a moderate band that keeps gradients informative.

\subsection{Global gain modulation and integration}

The layer wise reaction diffusion mechanism balances activity across depth on a fast time scale but it does not determine the global firing regime. We therefore introduce a scalar gain $g_{e}$ that is shared across units and updated once per epoch $e$. This gain multiplies all layer wise scales and coherently shifts the effective thresholds of the network.

The update is driven by two aggregated signals. The first measures validation progress by comparing the current validation loss $L^{\text{val}}_{e}$ to the previous epoch:
$$
p_{e}=\frac{L^{\text{val}}_{e-1}-L^{\text{val}}_{e}}{\max\!\left(L^{\text{val}}_{e-1},\,\epsilon\right)}.
$$
The second is the network activity energy, computed as the epoch average of EMA smoothed per unit rates:
$$
E_{e}=\frac{1}{L\hat{T}}\sum_{u=1}^{L}\sum_{\tau=1}^{\hat{T}}\tilde{r}^{u}_{e,\tau}.
$$
Both signals are smoothed across epochs. When progress stalls while energy is high, the controller increases $g_{e}$ to suppress spikes; when progress improves while energy is low, it decreases $g_{e}$ to permit more spikes. The update is multiplicative, positivity preserving, and limited by a small trust region, so $g_{e}$ stabilizes near a point where further changes no longer improve validation loss and the energy remains close to its trend.

At runtime the simulator combines the current global gain with the inner loop scales to set thresholds for all controlled units. In step mode $s$ thresholds are written at each physical timestep; in rate mode $m$ they are written once per virtual micro step. This uniform integration steers the SNN toward an effective operating band while keeping the training graph unchanged.

\section{Experiment}
\label{sec:formatting}

\begin{table*}[ht]
  \centering
  \setlength{\tabcolsep}{4.5pt}  
  \renewcommand{\arraystretch}{1.25}  
  
  {\setlength{\arrayrulewidth}{0.3pt}
  \begin{tabular}{c|c|c|c|ccc|ccc}
    \hline
    \multirow[c]{3}{*}{\shortstack{Training\\Paradigm}} &
    \multirow[c]{3}{*}{\shortstack{Step\\Mode}} &
    \multirow[c]{3}{*}{Backbone} &
    \multirow[c]{3}{*}{Method} &
    \multicolumn{6}{c}{$\Delta$ TOP-1 $Acc_{ID}$(\%)} \\
    \cline{5-10}
     & & & &
      \multicolumn{3}{c|}{event} & \multicolumn{3}{c}{rgb} \\
    \cline{5-7}\cline{8-10}
     & & & &
      T=6 & T=8 & T=10 &
      T=6 & T=8 & T=10 \\
    \hline

\multirow{12}{*}{BPTT} & \multirow{6}{*}{s} & ResNet-18 & AHSAR
            & +0.64 & +0.71 & +0.66 & +0.98 & +0.32 & +1.94 \\
         &   & ResNet-18 & AHSAR(no RT)
            & +0.26 & +0.42 & +0.32 & +0.44 & +0.15 & +1.51 \\
         &   & ResNet-18 & AHSAR(no DH)
            & +0.22 & +0.42 & +0.17 & +0.17 & +0.04 & +1.24 \\ \cline{3-10}
         &   & VGGNet-13 & AHSAR
            & +0.51 & +0.62 & +0.55 & +1.24 & +1.56 & +1.31 \\
         &   & VGGNet-13 & AHSAR(no RT)
            & +0.14 & +0.41 & +0.22 & +0.26 & +0.63 & +0.98 \\
         &   & VGGNet-13 & AHSAR(no DH)
            & +0.14 & +0.07 & +0.19 & +0.02 & +0.15 & +0.45 \\ \cline{2-10}

         & \multirow{6}{*}{m} & ResNet-18 & AHSAR
            & +1.11 & +1.06 & +1.03 & +1.65 & +0.51 & +0.77 \\
         &   & ResNet-18 & AHSAR(no RT)
            & +0.54 & +0.68 & +0.70 & +1.23 & +0.22 & +0.31 \\
         &   & ResNet-18 & AHSAR(no DH)
            & +0.40 & +0.60 & +0.51 & +0.92 & +0.26 & +0.46 \\ \cline{3-10}
         &   & VGGNet-13 & AHSAR
            & +0.43 & +0.81 & +0.30 & +1.06 & +1.85 & +1.33 \\
         &   & VGGNet-13 & AHSAR(no RT)
            & +0.20 & +0.42 & +0.12 & +0.61 & +1.42 & +0.79 \\
         &   & VGGNet-13 & AHSAR(no DH)
            & +0.14 & +0.21 & +0.12 & +0.60 & +1.38 & +0.51 \\ \hline

\multirow{12}{*}{Rate \cite{yu2024advancing}} & \multirow{6}{*}{s} & ResNet-18 & AHSAR
            & +1.15 & +1.21 & +1.22 & +1.37 & +1.95 & +1.45 \\
         &   & ResNet-18 & AHSAR(no RT)
            & +0.90 & +0.81 & +0.94 & +1.01 & +1.62 & +1.20 \\
         &   & ResNet-18 & AHSAR(no DH)
            & +0.74 & +0.70 & +0.81 & +0.45 & +1.10 & +1.00 \\ \cline{3-10}
         &   & VGGNet-13 & AHSAR
            & +0.81 & +0.86 & +0.76 & +1.12 & +1.73 & +2.35 \\
         &   & VGGNet-13 & AHSAR(no RT)
            & +0.32 & +0.45 & +0.21 & +0.66 & +1.55 & +1.87 \\
         &   & VGGNet-13 & AHSAR(no DH)
            & +0.33 & +0.31 & +0.50 & +0.45 & +1.35 & +1.74 \\ \cline{2-10}

         & \multirow{6}{*}{m} & ResNet-18 & AHSAR
            & +1.30 & +1.32 & +1.10 & +1.86 & +2.03 & +1.88 \\
         &   & ResNet-18 & AHSAR(no RT)
            & +0.96 & +0.84 & +0.94 & +1.42 & +1.51 & +1.49 \\
         &   & ResNet-18 & AHSAR(no DH)
            & +0.71 & +0.43 & +0.73 & +1.17 & +1.26 & +1.14 \\ \cline{3-10}
         &   & VGGNet-13 & AHSAR
            & +0.64 & +0.70 & +0.41 & +1.63 & +1.98 & +1.84 \\
         &   & VGGNet-13 & AHSAR(no RT)
            & +0.47 & +0.28 & +0.22 & +1.21 & +1.51 & +1.59 \\
         &   & VGGNet-13 & AHSAR(no DH)
            & +0.33 & +0.12 & +0.20 & +1.06 & +1.19 & +1.23 \\ \hline
  \end{tabular}}
  \caption{Comparison of incorporating the AHSAR method under different training paradigms (trained for 50 epochs).}
  \vspace{-5pt}
  \label{tab1}
\end{table*}

\subsection{Setup}
We systematically evaluate VGGNet and ResNet backbones on CIFAR-$10$ and CIFAR-$10$-DVS, covering both rate-based propagation and Backpropagation Through Time. Unless noted otherwise, we use a unified training setup with Adam, learning rate $10^{-3}$, weight decay $5\times10^{-4}$, a cosine learning rate schedule, and batch size $128$. We run single step and multi step temporal rollouts with $T$ equal to $6$, $8$, or $10$. Our AHSAR module attaches to selected LIF layers to learn per layer threshold scaling and writes back an EMA smoothed scale at evaluation with momentum $0.1$.

\label{sec:formatting}
\begin{table*}[ht]
  \centering
  \setlength{\tabcolsep}{4.5pt}
  \renewcommand{\arraystretch}{1.25}
  
  {\setlength{\arrayrulewidth}{0.3pt}
  \begin{tabular}{c|c|c|c|ccc|ccc}
    \hline
    \multirow[c]{3}{*}{\shortstack{Training\\Paradigm}} &
    \multirow[c]{3}{*}{\shortstack{Step\\Mode}} &
    \multirow[c]{3}{*}{Backbone} &
    \multirow[c]{3}{*}{Method} &
    \multicolumn{6}{c}{$\Delta$ TOP-1 $Acc_{OOD}$(\%)} \\
    \cline{5-10}
     & & & &
      \multicolumn{3}{c|}{event} & \multicolumn{3}{c}{rgb} \\
    \cline{5-7}\cline{8-10}
     & & & &
      T=6 & T=8 & T=10 &
      T=6 & T=8 & T=10 \\
    \hline

\multirow{12}{*}{BPTT} & \multirow{6}{*}{s} & ResNet-18 & AHSAR
            & +0.68 & +0.59 & +0.47 & +0.65 & +0.42 & +0.53 \\
         &   & ResNet-18 & AHSAR(no RT)
            & +0.32 & +0.41 & +0.26 & +0.38 & +0.25 & +0.31 \\
         &   & ResNet-18 & AHSAR(no DH)
            & +0.28 & +0.35 & +0.19 & +0.29 & +0.17 & +0.24 \\ \cline{3-10}
         &   & VGGNet-13 & AHSAR
            & +0.61 & +0.54 & +0.49 & +0.58 & +0.52 & +0.55 \\
         &   & VGGNet-13 & AHSAR(no RT)
            & +0.37 & +0.32 & +0.28 & +0.35 & +0.31 & +0.33 \\
         &   & VGGNet-13 & AHSAR(no DH)
            & +0.25 & +0.23 & +0.21 & +0.27 & +0.24 & +0.22 \\ \cline{2-10}

         & \multirow{6}{*}{m} & ResNet-18 & AHSAR
            & +0.63 & +0.57 & +0.52 & +0.61 & +0.48 & +0.54 \\
         &   & ResNet-18 & AHSAR(no RT)
            & +0.39 & +0.35 & +0.31 & +0.37 & +0.29 & +0.32 \\
         &   & ResNet-18 & AHSAR(no DH)
            & +0.28 & +0.26 & +0.23 & +0.25 & +0.21 & +0.24 \\ \cline{3-10}
         &   & VGGNet-13 & AHSAR
            & +0.59 & +0.53 & +0.46 & +0.57 & +0.49 & +0.51 \\
         &   & VGGNet-13 & AHSAR(no RT)
            & +0.36 & +0.32 & +0.28 & +0.34 & +0.30 & +0.29 \\
         &   & VGGNet-13 & AHSAR(no DH)
            & +0.24 & +0.22 & +0.19 & +0.23 & +0.20 & +0.18 \\ \hline

\multirow{12}{*}{Rate \cite{yu2024advancing}} & \multirow{6}{*}{s} & ResNet-18 & AHSAR
            & +0.67 & +0.62 & +0.55 & +0.64 & +0.58 & +0.60 \\
         &   & ResNet-18 & AHSAR(no RT)
            & +0.41 & +0.38 & +0.33 & +0.39 & +0.35 & +0.37 \\
         &   & ResNet-18 & AHSAR(no DH)
            & +0.29 & +0.27 & +0.24 & +0.28 & +0.25 & +0.26 \\ \cline{3-10}
         &   & VGGNet-13 & AHSAR
            & +0.65 & +0.59 & +0.51 & +0.63 & +0.56 & +0.58 \\
         &   & VGGNet-13 & AHSAR(no RT)
            & +0.38 & +0.35 & +0.30 & +0.36 & +0.33 & +0.34 \\
         &   & VGGNet-13 & AHSAR(no DH)
            & +0.26 & +0.24 & +0.21 & +0.25 & +0.23 & +0.22 \\ \cline{2-10}

         & \multirow{6}{*}{m} & ResNet-18 & AHSAR
            & +0.70 & +0.64 & +0.57 & +0.68 & +0.61 & +0.63 \\
         &   & ResNet-18 & AHSAR(no RT)
            & +0.43 & +0.39 & +0.35 & +0.41 & +0.37 & +0.38 \\
         &   & ResNet-18 & AHSAR(no DH)
            & +0.31 & +0.28 & +0.25 & +0.30 & +0.26 & +0.27 \\ \cline{3-10}
         &   & VGGNet-13 & AHSAR
            & +0.66 & +0.60 & +0.53 & +0.65 & +0.58 & +0.59 \\
         &   & VGGNet-13 & AHSAR(no RT)
            & +0.40 & +0.36 & +0.32 & +0.38 & +0.34 & +0.35 \\
         &   & VGGNet-13 & AHSAR(no DH)
            & +0.27 & +0.25 & +0.22 & +0.26 & +0.23 & +0.24 \\ \hline
  \end{tabular}}
  \caption{Comparison of incorporating the AHSAR method under different training paradigms (trained for 50 epochs). Here, RT denotes the relaxation term, while DH represents diffusion homeostasis and $\Delta$ denotes the difference with respect to the baseline..}
  \vspace{-5pt}
  \label{tab3}
\end{table*}

\begin{table}[t]
  \centering
  \small  
  \setlength{\tabcolsep}{2.5pt}  
  \renewcommand{\arraystretch}{1.1}  
  \begin{tabular}{l c c l c c}
    \toprule
    Model & Type & T & Arch. & Epochs & Acc.(\%) \\
    \midrule
    QCFS \cite{bu2023optimal}    & ANN2SNN & 8  & ResNet-18 & 300 & 94.82 \\
    IM-Loss \cite{guo2022loss}    & SNN & 20 & VGG-9       & 300 & 87.62 \\
    DeepEISNN \cite{liu2025training}  & SNN & 64 & VGG-8       & 300 & 87.73 \\
    IM-Loss \cite{guo2022loss}   & SNN & 4  & CIFARNet    & 300 & 90.90 \\
    DSR \cite{meng2022training}  & SNN & 4  & PA-ResNet-18 & 300 & 95.10 \\
    SSF \cite{wang2023ssf}        & SNN & 4  & PA-ResNet-18 & 300 & 94.90 \\
    \multirow{2}{*}{\textbf{AHSAR (Ours)}} & \multirow{2}{*}{\textbf{SNN}} & \multirow{2}{*}{\textbf{4}} & 
    \multirow{2}{*}{\textbf{ResNet-18}} & \textbf{50}  & \textbf{92.77} \\
    & & & & \textbf{300}  & \textbf{95.92} \\
    \bottomrule
  \end{tabular}%
  \caption{Comparison with state-of-the-art models, where PA-ResNet-18 denotes PreAct-ResNet-18.}
  \label{tab2}
\end{table}

\subsection{Quantitative Results}
\paragraph{On In-Distribution Data.} To rigorously assess the contribution of our approach, we evaluate it on a suite of widely used strong baselines and integrate it as a standalone module that preserves the original network architecture and training paradigm. We conduct controlled comparisons across training regimes and key hyperparameters, training each configuration for 50 epochs, with results summarized in Table \ref{tab1}. We further apply the proposed method to the Rate-based Propagation baseline and compare it against the current state-of-the-art methods reported in Table \ref{tab2}. The results consistently show that the approach functions as a plug and play component that improves optimization stability, accelerates convergence, and enhances final performance without altering the underlying training paradigm.

\paragraph{On Out-of-Distribution Data.}Following the experimental setup for in-distribution (ID) data accuracy evaluation, we conducted an identically configured experiment for out-of-distribution (OOD) data accuracy evaluation, with results shown in Table \ref{tab3}. 
Additionally, to more comprehensively evaluate the model's performance in OOD detection, we adopted predictive entropy as the detection score and reported the AUROC and FPR@95TPR. Intuitively, predictive entropy reflects the uncertainty of the model's predictions—a higher predictive entropy (indicating greater model uncertainty) suggests that the sample is more likely to be OOD. Specifically, the AUROC provides a threshold-free overall measure of separability, which can be interpreted as the probability that a randomly chosen OOD sample receives a higher detection score than an ID sample; a higher AUROC indicates better discrimination between ID and OOD samples. Conversely, FPR@95TPR denotes the false positive rate of ID samples when the recall rate for OOD samples is fixed at 95\%; a lower value indicates a stronger ability to control false positives while maintaining high recall. Predictive entropy is used to quantify the overall uncertainty of the predictive distribution. The AUROC and FPR@95TPR results are shown in Table \ref{tab5}.

\begin{table*}[ht]
  \centering
  \setlength{\tabcolsep}{4.5pt}
  \renewcommand{\arraystretch}{1.25}
  
  {\setlength{\arrayrulewidth}{0.3pt}
  \begin{tabular}{c|c|c|ccc|ccc|ccc|ccc}
    \hline
    \multirow[c]{3}{*}{\shortstack{Training\\Paradigm}} &
    \multirow[c]{3}{*}{\shortstack{Step\\Mode}} &
    \multirow[c]{3}{*}{Backbone} &
    \multicolumn{6}{c|}{$\Delta$ $AUROC_{Entropy}$(\%)} &
    \multicolumn{6}{c}{$\Delta$ $FPR95\%TPR_{Entropy}$(\%)} \\
    \cline{4-9}\cline{10-15}
     & & &
      \multicolumn{3}{c|}{event} & \multicolumn{3}{c|}{rgb} &
      \multicolumn{3}{c|}{event} & \multicolumn{3}{c}{rgb} \\
    \cline{4-6}\cline{7-9}\cline{10-12}\cline{13-15}
     & & &
      T=6 & T=8 & T=10 &
      T=6 & T=8 & T=10 &
      T=6 & T=8 & T=10 &
      T=6 & T=8 & T=10 \\
    \hline

\multirow{4}{*}{BPTT} & \multirow{2}{*}{s} & ResNet-18 
            & +6 & +11 & +7 & +26 & +27 & +24 & -4 & -6 & -8 & -15 & -17 & -19 \\
         \cline{3-15}
         &   & VGGNet-13 
            & +7 & +11 & +8 & +21 & +22 & +20 & -5 & -7 & -9 & -16 & -18 & -20 \\
         \cline{2-15}

         & \multirow{2}{*}{m} & ResNet-18 
            & +8 & +7 & +4 & +17 & +20 & +25 & -3 & -5 & -10 & -14 & -19 & -21 \\
         \cline{3-15}
         &   & VGGNet-13 
            & +6 & +2 & +2 & +11 & +26 & +18 & -6 & -8 & -11 & -17 & -20 & -22 \\
         \hline

\multirow{4}{*}{Rate \cite{yu2024advancing}} & \multirow{2}{*}{s} & ResNet-18
            & +8 & +1 & +3 & +16 & +26 & +20 & -4 & -7 & -9 & -15 & -18 & -23 \\
         \cline{3-15}
         &   & VGGNet-13 
            & +7 & +6 & +17 & +21 & +17 & +25 & -5 & -8 & -12 & -16 & -19 & -24 \\
         \cline{2-15}

         & \multirow{2}{*}{m} & ResNet-18 
            & +8 & +6 & +4 & +18 & +21 & +17 & -3 & -6 & -10 & -13 & -20 & -25 \\
         \cline{3-15}
         &   & VGGNet-13 
            & +8 & +4 & +7 & +15 & +17 & +15 & -4 & -9 & -11 & -14 & -21 & -26 \\
         \hline
  \end{tabular}}
  \caption{Comparison of incorporating the AHSAR method under different training paradigms (trained for 50 epochs), where $\Delta$ denotes the difference with respect to the baseline.}
  \vspace{-5pt}
  \label{tab_combined}
\end{table*}

\subsection{Qualitative Results}

\begin{figure}[t]
  \centering
  \includegraphics[width=0.95\linewidth]{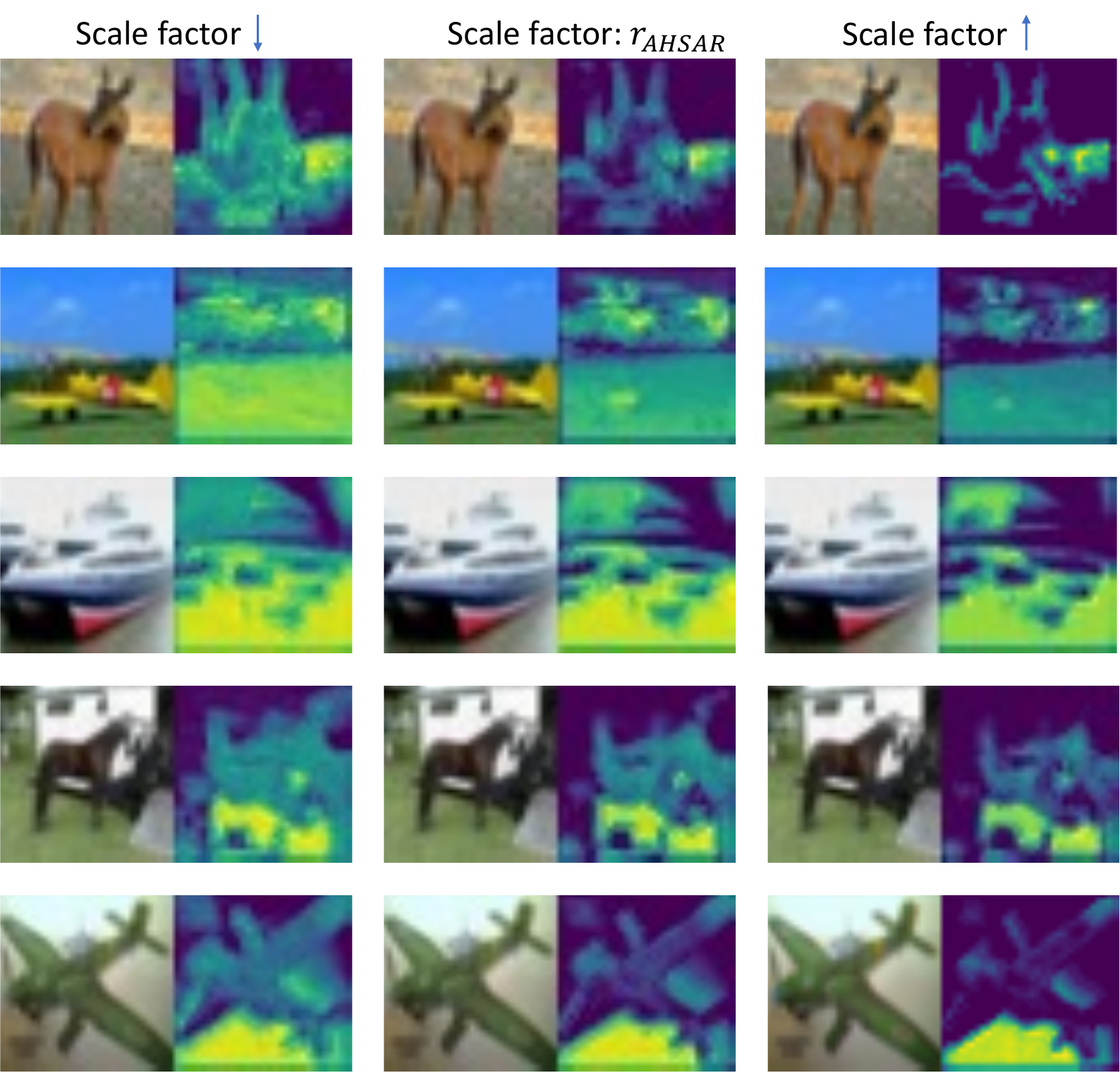}
  \vspace{-5pt}
  \caption{Spiking activity under different firing thresholds, where the "Scale factor" represents the scaling factor applied to the original firing threshold. $r_{AHSAR}$ denotes the scaling factor obtained using the AHSAR method.}
  \vspace{-15pt}
  \label{fig4}
\end{figure}
To make the effect of our method on spiking dynamics explicit, Figure \ref{fig4} visualizes the layerwise spike rasters together with the corresponding firing rate maps. When the scaling factor is small, the effective threshold decreases and neurons fire easily. The firing rate increases, random spikes spread across large regions, and background areas exhibit pronounced noise. When the scaling factor is large, the effective threshold increases and spiking activity is suppressed. Many informative events fail to trigger spikes, which reduces the amount of captured information and weakens the representation. These observations indicate that appropriate scaling is essential to balance noise suppression and information preservation. Our strategy maintains the firing rate within a stable range, producing clear responses at salient structures while avoiding the accumulation of background noise.

\subsection{Ablation Studies}
To examine the contribution of each component in greater detail, we conducted a systematic ablation study. All experiments used identical data splits and training settings. We performed ablations on the AHSAR method by removing the inter-layer diffusion term and the relaxation term to observe the impact of these components on model training convergence. The results are shown in Table 1. It can be observed that across different training paradigms and data modalities, our method enhances the speed of model convergence.

\section{Conclusion}
\label{sec:formatting}

AHSAR offers a lightweight, plug-and-play solution to stabilize and accelerate SNN training. It maintains layer-wise spiking activity within a productive band through dual-timescale control—fast reaction-diffusion balancing and slow global gain modulation—without adding parameters or changing model architecture. Experiments confirm faster convergence and higher accuracy across architectures, datasets, and training paradigms, supporting efficient deployment in low-power vision systems.



\begin{thebibliography}{37}
\providecommand{\natexlab}[1]{#1}
\providecommand{\url}[1]{\texttt{#1}}
\expandafter\ifx\csname urlstyle\endcsname\relax
  \providecommand{\doi}[1]{doi: #1}\else
  \providecommand{\doi}{doi: \begingroup \urlstyle{rm}\Url}\fi

\bibitem[Abbott and Nelson(2000)]{abbott2000synaptic}
Larry~F Abbott and Sacha~B Nelson.
\newblock Synaptic plasticity: taming the beast.
\newblock \emph{Nature neuroscience}, 3\penalty0 (11):\penalty0 1178--1183, 2000.

\bibitem[Amir et~al.(2017)Amir, Taba, Berg, Melano, McKinstry, Di~Nolfo, Nayak, Andreopoulos, Garreau, Mendoza, et~al.]{amir2017low}
Arnon Amir, Brian Taba, David Berg, Timothy Melano, Jeffrey McKinstry, Carmelo Di~Nolfo, Tapan Nayak, Alexander Andreopoulos, Guillaume Garreau, Marcela Mendoza, et~al.
\newblock A low power, fully event-based gesture recognition system.
\newblock In \emph{Proceedings of the IEEE conference on computer vision and pattern recognition}, pages 7243--7252, 2017.

\bibitem[Aston-Jones and Cohen(2005)]{aston2005integrative}
Gary Aston-Jones and Jonathan~D Cohen.
\newblock An integrative theory of locus coeruleus-norepinephrine function: adaptive gain and optimal performance.
\newblock \emph{Annu. Rev. Neurosci.}, 28\penalty0 (1):\penalty0 403--450, 2005.

\bibitem[Attwell and Laughlin(2001)]{attwell2001energy}
David Attwell and Simon~B. Laughlin.
\newblock An energy budget for signaling in the grey matter of the brain.
\newblock \emph{Journal of Cerebral Blood Flow and Metabolism}, 21\penalty0 (10):\penalty0 1133--1145, 2001.

\bibitem[Bu et~al.(2022)Bu, Ding, Yu, and Huang]{bu2022optimized}
Tong Bu, Jianhao Ding, Zhaofei Yu, and Tiejun Huang.
\newblock Optimized potential initialization for low-latency spiking neural networks.
\newblock In \emph{Proceedings of the AAAI conference on artificial intelligence}, pages 11--20, 2022.

\bibitem[Bu et~al.(2023)Bu, Fang, Ding, Dai, Yu, and Huang]{bu2023optimal}
Tong Bu, Wei Fang, Jianhao Ding, PengLin Dai, Zhaofei Yu, and Tiejun Huang.
\newblock Optimal ann-snn conversion for high-accuracy and ultra-low-latency spiking neural networks.
\newblock \emph{arXiv preprint arXiv:2303.04347}, 2023.

\bibitem[Cao et~al.(2015)Cao, Chen, and Khosla]{cao2015spiking}
Yongqiang Cao, Yang Chen, and Deepak Khosla.
\newblock Spiking deep convolutional neural networks for energy-efficient object recognition.
\newblock \emph{International Journal of Computer Vision}, 113\penalty0 (1):\penalty0 54--66, 2015.

\bibitem[Chen et~al.(2023)Chen, Peng, Li, and Tian]{chen2023training}
Guangyao Chen, Peixi Peng, Guoqi Li, and Yonghong Tian.
\newblock Training full spike neural networks via auxiliary accumulation pathway.
\newblock \emph{arXiv preprint arXiv:2301.11929}, 2023.

\bibitem[Diehl et~al.(2015)Diehl, Neil, Binas, Cook, Liu, and Pfeiffer]{diehl2015fast}
Peter~U Diehl, Daniel Neil, Jonathan Binas, Matthew Cook, Shih-Chii Liu, and Michael Pfeiffer.
\newblock Fast-classifying, high-accuracy spiking deep networks through weight and threshold balancing.
\newblock In \emph{2015 International joint conference on neural networks (IJCNN)}, pages 1--8. ieee, 2015.

\bibitem[Ding et~al.(2022)Ding, Dong, Heide, Ding, Zhou, Yin, and Yang]{ding2022biologically}
Jianchuan Ding, Bo Dong, Felix Heide, Yufei Ding, Yunduo Zhou, Baocai Yin, and Xin Yang.
\newblock Biologically inspired dynamic thresholds for spiking neural networks.
\newblock \emph{Advances in neural information processing systems}, 35:\penalty0 6090--6103, 2022.

\bibitem[Duan et~al.(2022)Duan, Ding, Chen, Yu, and Huang]{duan2022temporal}
Chaoteng Duan, Jianhao Ding, Shiyan Chen, Zhaofei Yu, and Tiejun Huang.
\newblock Temporal effective batch normalization in spiking neural networks.
\newblock In \emph{Advances in Neural Information Processing Systems}, pages 34377--34390, 2022.

\bibitem[Dunwiddie and Masino(2001)]{dunwiddie2001role}
Thomas~V Dunwiddie and Susan~A Masino.
\newblock The role and regulation of adenosine in the central nervous system.
\newblock \emph{Annual review of neuroscience}, 24\penalty0 (1):\penalty0 31--55, 2001.

\bibitem[Guo et~al.(2022)Guo, Chen, Zhang, Liu, Wang, Huang, and Ma]{guo2022loss}
Yufei Guo, Yuanpei Chen, Liwen Zhang, Xiaode Liu, Yinglei Wang, Xuhui Huang, and Zhe Ma.
\newblock Im-loss: information maximization loss for spiking neural networks.
\newblock \emph{Advances in Neural Information Processing Systems}, 35:\penalty0 156--166, 2022.

\bibitem[Guo et~al.(2024)Guo, Peng, Liu, Chen, Zhang, Tong, Jie, and Ma]{guo2024enof}
Yufei Guo, Weihang Peng, Xiaode Liu, Yuanpei Chen, Yuhan Zhang, Xin Tong, Zhou Jie, and Zhe Ma.
\newblock Enof-snn: Training accurate spiking neural networks via enhancing the output feature.
\newblock \emph{Advances in Neural Information Processing Systems}, 37:\penalty0 51708--51726, 2024.

\bibitem[Ho and Chang(2021)]{ho2021tcl}
Nguyen-Dong Ho and Ik-Joon Chang.
\newblock Tcl: an ann-to-snn conversion with trainable clipping layers.
\newblock In \emph{2021 58th ACM/IEEE design automation conference (DAC)}, pages 793--798. IEEE, 2021.

\bibitem[Jiang et~al.(2023)Jiang, Anumasa, De~Masi, Xiong, and Gu]{jiang2023unified}
Haiyan Jiang, Srinivas Anumasa, Giulia De~Masi, Huan Xiong, and Bin Gu.
\newblock A unified optimization framework of ann-snn conversion: towards optimal mapping from activation values to firing rates.
\newblock In \emph{International Conference on Machine Learning}, pages 14945--14974. PMLR, 2023.

\bibitem[Kim and Panda(2021)]{kim2021revisiting}
Youngeun Kim and Priyadarshini Panda.
\newblock Revisiting batch normalization for training low-latency deep spiking neural networks from scratch.
\newblock \emph{Frontiers in Neuroscience}, 15:\penalty0 773954, 2021.

\bibitem[Lee et~al.(2020)Lee, Sarwar, Panda, Srinivasan, and Roy]{lee2020enabling}
Chankyu Lee, Syed~Shakib Sarwar, Priyadarshini Panda, Gopalakrishnan Srinivasan, and Kaushik Roy.
\newblock Enabling spike-based backpropagation for training deep neural network architectures.
\newblock \emph{Frontiers in neuroscience}, 14:\penalty0 497482, 2020.

\bibitem[Lee et~al.(2016)Lee, Delbruck, and Pfeiffer]{lee2016training}
Jun~Haeng Lee, Tobi Delbruck, and Michael Pfeiffer.
\newblock Training deep spiking neural networks using backpropagation.
\newblock \emph{Frontiers in neuroscience}, 10:\penalty0 508, 2016.

\bibitem[Lennie(2003)]{lennie2003cost}
Peter Lennie.
\newblock The cost of cortical computation.
\newblock \emph{Current Biology}, 13\penalty0 (6):\penalty0 493--497, 2003.

\bibitem[Li et~al.(2021)Li, Guo, Zhang, Deng, Hai, and Gu]{li2021differentiable}
Yuhang Li, Yufei Guo, Shanghang Zhang, Shikuang Deng, Yongqing Hai, and Shi Gu.
\newblock Differentiable spike: Rethinking gradient-descent for training spiking neural networks.
\newblock In \emph{Advances in Neural Information Processing Systems}, pages 23426--23439, 2021.

\bibitem[Liu et~al.(2025)Liu, Ding, and Yu]{liu2025training}
Peiyu Liu, Jianhao Ding, and Zhaofei Yu.
\newblock Training deep normalization-free spiking neural networks with lateral inhibition.
\newblock \emph{arXiv preprint arXiv:2509.23253}, 2025.

\bibitem[Marder and Goaillard(2006)]{marder2006variability}
Eve Marder and Jean-Marc Goaillard.
\newblock Variability, compensation and homeostasis in neuron and network function.
\newblock \emph{Nature Reviews Neuroscience}, 7\penalty0 (7):\penalty0 563--574, 2006.

\bibitem[Meng et~al.(2022)Meng, Xiao, Yan, Wang, Lin, and Luo]{meng2022training}
Qingyan Meng, Mingqing Xiao, Shen Yan, Yisen Wang, Zhouchen Lin, and Zhi-Quan Luo.
\newblock Training high-performance low-latency spiking neural networks by differentiation on spike representation.
\newblock In \emph{Proceedings of the IEEE/CVF conference on computer vision and pattern recognition}, pages 12444--12453, 2022.

\bibitem[Meng et~al.(2023)Meng, Xiao, Yan, Wang, Lin, and Luo]{meng2023towards}
Qingyan Meng, Mingqing Xiao, Shen Yan, Yisen Wang, Zhouchen Lin, and Zhi-Quan Luo.
\newblock Towards memory-and time-efficient backpropagation for training spiking neural networks.
\newblock In \emph{Proceedings of the IEEE/CVF international conference on computer vision}, pages 6166--6176, 2023.

\bibitem[Neftci et~al.(2019)Neftci, Mostafa, and Zenke]{neftci2019surrogate}
Emre~O. Neftci, Hesham Mostafa, and Friedemann Zenke.
\newblock Surrogate gradient learning in spiking neural networks: Bringing the power of gradient-based optimization to spiking neural networks.
\newblock \emph{IEEE Signal Processing Magazine}, 36\penalty0 (6):\penalty0 51--63, 2019.

\bibitem[Orchard et~al.(2015)Orchard, Jayawant, Cohen, and Thakor]{orchard2015converting}
Garrick Orchard, Ajinkya Jayawant, Gregory~K Cohen, and Nitish Thakor.
\newblock Converting static image datasets to spiking neuromorphic datasets using saccades.
\newblock \emph{Frontiers in neuroscience}, 9:\penalty0 437, 2015.

\bibitem[Picciotto et~al.(2012)Picciotto, Higley, and Mineur]{picciotto2012acetylcholine}
Marina~R Picciotto, Michael~J Higley, and Yann~S Mineur.
\newblock Acetylcholine as a neuromodulator: cholinergic signaling shapes nervous system function and behavior.
\newblock \emph{Neuron}, 76\penalty0 (1):\penalty0 116--129, 2012.

\bibitem[Roy et~al.(2019)Roy, Jaiswal, and Panda]{roy2019towards}
Kaushik Roy, Akhilesh Jaiswal, and Priyadarshini Panda.
\newblock Towards spike-based machine intelligence with neuromorphic computing.
\newblock \emph{Nature}, 575\penalty0 (7784):\penalty0 607--617, 2019.

\bibitem[Ruggiero et~al.(2021)Ruggiero, Katsenelson, and Slutsky]{ruggiero2021mitochondria}
Antonella Ruggiero, Maxim Katsenelson, and Inna Slutsky.
\newblock Mitochondria: new players in homeostatic regulation of firing rate set points.
\newblock \emph{Trends in Neurosciences}, 44\penalty0 (8):\penalty0 605--618, 2021.

\bibitem[Shrestha and Orchard(2018)]{shrestha2018slayer}
Sumit~B Shrestha and Garrick Orchard.
\newblock Slayer: Spike layer error reassignment in time.
\newblock \emph{Advances in neural information processing systems}, 31, 2018.

\bibitem[Turrigiano(2008)]{turrigiano2008self}
Gina~G. Turrigiano.
\newblock The self-tuning neuron: Synaptic scaling of excitatory synapses.
\newblock \emph{Cell}, 135\penalty0 (3):\penalty0 422--435, 2008.

\bibitem[Turrigiano and Nelson(2004)]{turrigiano2004homeostatic}
Gina~G. Turrigiano and Sacha~B. Nelson.
\newblock Homeostatic plasticity in the developing nervous system.
\newblock \emph{Nature Reviews Neuroscience}, 5\penalty0 (2):\penalty0 97--107, 2004.

\bibitem[Vogels et~al.(2011)Vogels, Sprekeler, Zenke, Clopath, and Gerstner]{vogels2011inhibitory}
Tim~P. Vogels, Henning Sprekeler, Friedemann Zenke, Claudia Clopath, and Wulfram Gerstner.
\newblock Inhibitory plasticity balances excitation and inhibition in sensory pathways and memory networks.
\newblock \emph{Science}, 334\penalty0 (6062):\penalty0 1569--1573, 2011.

\bibitem[Wang et~al.(2023)Wang, Song, Wang, Xiao, Yang, Mei, and Zhang]{wang2023ssf}
Jingtao Wang, Zengjie Song, Yuxi Wang, Jun Xiao, Yuran Yang, Shuqi Mei, and Zhaoxiang Zhang.
\newblock Ssf: Accelerating training of spiking neural networks with stabilized spiking flow.
\newblock In \emph{Proceedings of the IEEE/CVF International Conference on Computer Vision}, pages 5982--5991, 2023.

\bibitem[Yu et~al.(2024)Yu, Liu, Wang, Li, and Wang]{yu2024advancing}
Chengting Yu, Lei Liu, Gaoang Wang, Erping Li, and Aili Wang.
\newblock Advancing training efficiency of deep spiking neural networks through rate-based backpropagation.
\newblock In \emph{Advances in Neural Information Processing Systems}, 2024.

\bibitem[Zenke et~al.(2017)Zenke, Gerstner, and Ganguli]{zenke2017temporal}
Friedemann Zenke, Wulfram Gerstner, and Surya Ganguli.
\newblock The temporal paradox of hebbian learning and homeostatic plasticity.
\newblock \emph{Current opinion in neurobiology}, 43:\penalty0 166--176, 2017.

\end{thebibliography}
\end{document}